\documentclass[]{interact}
\usepackage{tabu}
\usepackage{multirow}
\usepackage{cite}
\usepackage{adjustbox}

\usepackage{amsmath,amssymb,amsfonts}
\newcolumntype{C}[1]{>{\centering\let\newline\\\arraybackslash}m{#1}}
\usepackage{epstopdf}% To incorporate .eps illustrations using PDFLaTeX, etc.
\usepackage{subfigure}% Support for small, `sub' figures and tables
\newcommand{\ApproxSign}{\raise.17ex\hbox{$\scriptstyle\sim$}}
\usepackage{natbib}% Citation support using natbib.sty
\bibpunct[, ]{(}{)}{;}{a}{}{,}% Citation support using natbib.sty
\usepackage{tikz}
\newcommand*\circled[1]{\tikz[baseline=(char.base)]{
        \node[shape=circle,draw,minimum size=4mm, inner sep=0pt] (char)
        {\scriptsize\rule[-3pt]{0pt}{\dimexpr2ex+2pt}#1};}}

\begin{document}

%\articletype{ARTICLE TEMPLATE}

\title{ACLNet: An Attention and Clustering-based Cloud Segmentation Network}

\author{
\name{Dhruv Makwana\textsuperscript{1} and Subhrajit Nag\textsuperscript{2} and Onkar Susladkar\textsuperscript{3} and Gayatri Deshmukh\textsuperscript{3} and Sai~Chandra~Teja~R\textsuperscript{1} and Sparsh Mittal\textsuperscript{4} and C Krishna Mohan\textsuperscript{1,2}\thanks{CONTACT C Krishna Mohan. Email: ckm@cse.iith.ac.in. Dhruv and Subhrajit are co-first authors. This work was supported by WNI WxBunka Foundation, Japan. The computing systems used in this research were provided by Indian Institute of Technology, Roorkee, India under grant FIG-100874. }}
\affil{\textsuperscript{1}CKM~VIGIL~Pvt~Ltd, Hyderabad, India; \textsuperscript{2}CSE Department, IIT~Hyderabad, India;  \textsuperscript{3}CSE Department, Vishwakarma Institute of Information Technology, Pune, India; \textsuperscript{4}Mehta Family School of Data Science and Artificial Intelligence, IIT~Roorkee,~India}
}

\maketitle

% \begin{abstract}
% This template is for authors who are preparing a manuscript for a Taylor \& Francis journal using the \LaTeX\ document preparation system and the \texttt{interact} class file, which is available via selected journals' home pages on the Taylor \& Francis website.
% \end{abstract}
\begin{abstract}
 
We propose a novel deep learning model named ACLNet, for cloud segmentation from ground images. ACLNet uses both deep neural network and machine learning (ML) algorithm to extract complementary features. Specifically, it uses EfficientNet-B0 as the backbone, ``\`a trous spatial pyramid pooling'' (ASPP) to learn at multiple receptive fields, and ``global attention module'' (GAM) to extract fine-grained details from the image. ACLNet also uses \textit{k}-means clustering to extract cloud boundaries more precisely.  ACLNet is effective for both daytime and nighttime images. It provides lower error rate, higher recall and higher F1-score than state-of-art cloud segmentation models. The source-code of ACLNet is available here: https://github.com/ckmvigil/ACLNet.

\end{abstract}


\section{Introduction}\label{sec:introduction}
Cloud segmentation has several applications such as weather prediction, climate hazard prediction, and solar energy forecasting. A study of clouds can provide crucial clues about the hydrological scorecard of the atmosphere and impending climate hazards. Further, clouds hamper satellite cameras' views and affect solar energy generation plants. Driven by these factors, the detection of clouds has gained a lot of traction in recent years. However, an accurate segmentation of clouds has remained a challenge due to factors such as the non-rigid shape of clouds and variable lighting especially of nighttime images.

Previous works have performed cloud segmentation on images taken from satellites \citep{xie2017multilevel}, and those taken from the ground \citep{dev2016color, shi2020cloudu, xie2020segcloud}. 
We review those works that analyze cloud images taken from ground. 
Previous works use color information to distinguish cloud (blue) from the sky (white).  For example, in the technique of  \cite{long2015fully}, the cloud area is detected by setting a threshold value for the ratio of the pixel values of red and blue. However, this method assumes that the colors of the sky and clouds in the daytime can be clearly seen. But this is not effective for images with a small difference in color between the sky and clouds, such as images at night, dawn, or dusk.

\cite{shi2019diurnal} propose a VGG16-based fully-convolutional network. They incorporate ``histogram equalization'' and ``skip connections'' to achieve effective segmentation. 
\cite{xie2020segcloud} use an encoder-decoder architecture. They modify the VGG16 network by replacing the fully connected layers with the decoder network. 
``CloudU-Net'' \citep{shi2020cloudu} is inspired from U-Net and it uses ``dilated convolutions'' and ``fully connected conditional random field (CRF)''. It also uses a lookahead optimizer for faster model convergence.  ``CloudSegNet'' \citep{dev2019cloudsegnet} is an encoder-decoder architecture and has been evaluated on both daytime and nighttime images. Table \ref{tab:recentWorks} shows the key characteristics of a few related works. 
 
\begin{table}[htbp]\footnotesize
  \centering
  \caption{Key characteristics of methods used in recent studies (SLIC = simple linear iterative clustering, TCDD  = TJNU  Cloud Detection Database, TLCDD = TJNU Large-scale Cloud Detection Database)}
    \begin{tabular}{|C{3cm}|C{8cm}|C{2cm}|}
    \hline
          & Architecture and highlights & Dataset \\
    \hline
    CloudSegNet \citep{dev2019cloudsegnet}  & Encoder-decoder CNN with Conv and DeConv layers & SWIMSEG, SWINSEG \\
    \hline
    %Dev et al. 
    \cite{dev2014systematic} & Fuzzy c-means clustering      & HYTA \\
    \hline
    \cite{dev2017nighttime} & SLIC for superpixel generation and \textit{k}-means clustering       & SWINSEG \\
    \hline
    \cite{xie2017multilevel} & SLIC for superpixel generation and CNN for classifying superpixels into thick/thin-cloud and sky. CNN has 2 branches to extract features at two scales & Quickbird satellite, Google map, internet images \\
    \hline
     CloudU-Net    \citep{shi2020cloudu}  & Encoder-decoder CNN based on U-Net. Fully-connected CRF layers, dilated Conv & GDNCI \\
    \hline
     MACNN \citep{zhang2021ground} & Encoder-decoder CNN with attention. Dilated Conv with different dilation values to capture multiscale information &  TCDD \\\hline
    CloudRaednet \citep{shi2022cloudraednet} & Residual attention-based encoder–decoder CNN.  ResNet50 as encoder; residual modules in decoder; Ranger optimizer & GDNCI \\ \hline
    DPNet \citep{zhang2022ground} &  Encoder-decoder CNN. Spatial pyramid pooling and then fusing the features by attention weights  &  TLCDD \\\hline
    \end{tabular}%
  \label{tab:recentWorks}%
\end{table}%
 
 Further, several previous works such as ``CRF as recurrent neural networks'' (CRF-RNN) have large computation and model-size overheads.

 In this paper, we propose a DNN for segmenting clouds from both daytime and nighttime sky images. ACLNet utilizes EfficientNet-B0 \citep{tan2019efficientnet} as the backbone. This allows ACLNet to learn better feature maps since  EfficientNet-B0 is a highly regularized model. Also, we use the ``\`a trous spatial pyramid pooling'' (ASPP) module \citep{chen2018encoder} to learn at multiple receptive fields. This helps in detecting clouds of different sizes.  Further, ACLNet uses two key novelties. First, we propose a ``global attention module'' (GAM) to focus on regions of interest, strengthen the learning of channels, and improve training efficiency. Second, we propose using \textit{k}-means clustering for detecting cloud boundaries. This helps in learning features that complement those learnt by the DNN. We combine this information with the DNN-extracted features for generating the final segmentation mask.

ACLNet has lower error rate, higher recall and higher F1-score than the state-of-art models, for daytime, nighttime, and day+night time images (corresponding to SWIMSEG, SWINSEG and SWINySEG datasets, respectively). The main contributions of the paper are: (1) An interpretable and accurate framework targeted for cloud segmentation from both daytime and nighttime images. (2) We introduce an ML algorithm in our DNN to bring their best together and more effectively segment the clouds.

\section{ACLNet: Proposed network}\label{sec:proposed}
\begin{figure*}[htbp]  \centering
\includegraphics [scale=0.50] {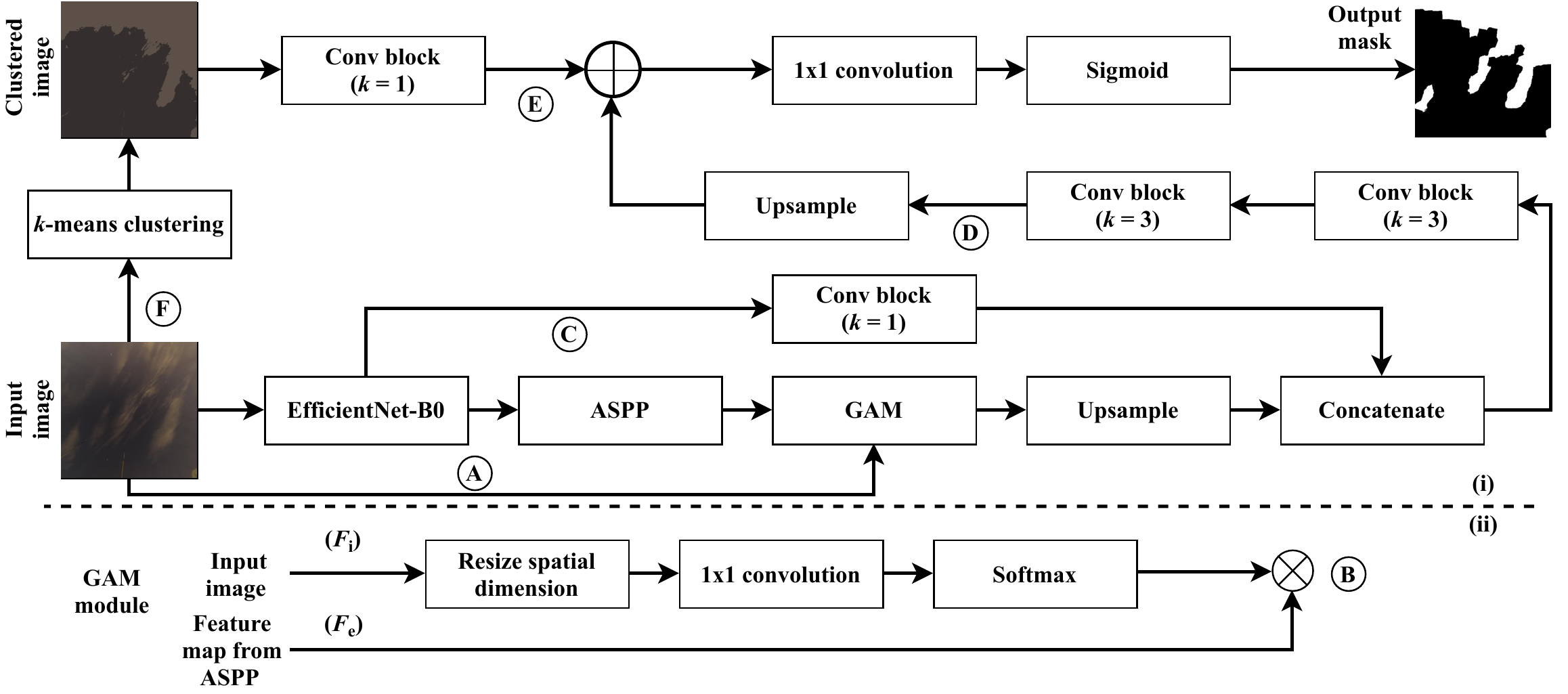}
\caption{ (i) ACLNet architecture (k denotes the kernel size) (ii) GAM module}
 \label{fig:ModelArchitecture}
  \end{figure*}

ACLNet is a novel network that outputs a cloud segmentation binary mask. Figure \ref{fig:ModelArchitecture} depicts the proposed ACLNet architecture. Here, a Conv block consists of a convolution layer followed by a batchnorm and ReLU activation. ACLNet uses EfficientNet-B0 \citep{tan2019efficientnet} as the backbone network.  EfficientNet-B0 is a highly regularised model. It has been obtained by performing a neural architecture search and jointly optimizing FLOPS, and accuracy. We now describe other components of ACLNet.

\subsection{ASPP Module}
Clouds vary widely in their size. To effectively detect clouds of various sizes, we seek to resample the features of a single scale. For this purpose, we use ASPP block \citep{chen2018encoder}. The input to ASPP is the downsampled feature map derived from the final layer of the backbone. ASPP uses parallel \`a trous convolution layers with different sampling rates, processes these features in different branches, and fuses them.  Specifically, ASPP has an average pooling layer with global information features, a Conv block with kernel size of $1 \times 1$ for extracting original scale features, and three  Conv blocks having kernel size of $3 \times 3$ with dilation rates of 6, 12, and 18, respectively. Thus, it uses multiple receptive fields. 
Finally, ASPP concatenates the feature maps and uses a Conv block with $1 \times 1$ kernel size to reduce the number of channels. 

While extracting multi-scale features, ASPP substantially reduces the information loss caused by numerous downsampling operations. It assesses convolutional features at several scales by applying \`a trous convolution with different dilation rates to image-level features. It reduces the number of parameters due to the use of three dilated convolutions. Furthermore, to maintain more location information, the sizes of distinct feature maps at different scales are kept the same.

\subsection{Global Attention Module (GAM)}\label{sec:GA<}

Motivation: We need an attention mechanism for two reasons. 
(1) The output from ASPP has little semantic information, but the input image provides rich semantic information that may be utilized to aid a low-level output from the ASPP feature in capturing semantic dependencies.  Therefore, improving the contextual information of the low-level output from the ASPP feature makes it easier to ensure effectual fusion.

(2) The input image contains rich contextual information. However, the contextual information provided by the ASPP output is inadequate for pixel-wise recognition.  To compensate for this loss of contextual information, one can concatenate the original input image with the ASPP output. However, concatenating  two incompatible features negatively impacts the semantic gaps between the ASPP output and the corresponding input image. We need to include additional semantic information into the ASPP output to fuse them effectively.

We introduce GAM to achieve above two objectives. GAM adaptively boosts location and semantic information by giving pixel-level and channel-level attention to the image.  The design of GAM is shown in Figure \ref{fig:ModelArchitecture}(ii).

Working of GAM: Let ASPP output be $F_{\text{e}} \in \text{R} ^ {H_1 \times W_1 \times C_1}$, where $H_1$ and $W_1$ are height and width of the feature map and $C_1$ is the number of channels. Similarly, let input image be $ F_\text{i} \in \text{R}^{H_2 \times W_2 \times C_2}$. In GAM, we proceed as follows:

(1) Spatially resize the image (refer \circled{A} in Figure \ref{fig:ModelArchitecture}) to match the spatial dimension of ASPP output. The image-dimension becomes $H_1 \times W_1 \times C_2$. (2) Use $1 \times 1$ convolution to capture channel dependencies for creating squeeze channel attention map of size $H_1 \times W_1 \times C_1$.
(3) Apply a softmax activation on the channel attention map to obtain the pixel-wise attention weights. This generates the attention map $A \in \text{R}^{H_1 \times W_1 \times C_1}$. 
(4) To create the final refined features, multiply the above attention map with ASPP output feature map $F_\text{e}$. This produces the output shown as \circled{B} in Figure \ref{fig:ModelArchitecture}.

The output of GAM is up-sampled to refine the features. We concatenate this upsampled feature map with the matching low-level feature map from the network backbone, which has the same spatial resolution. Before this concatenation, we pass the low-level feature map obtained from the backbone network (refer \circled{C} in Figure \ref{fig:ModelArchitecture}) through a Conv block with kernel size of $1 \times 1$ to reduce the number of channels. This is because the low-level features typically contain a large number of channels, which can outweigh the importance of the output from ASPP features. 

\subsection{Using clustered image}
Recent cloud-segmentation techniques exclusively use DNN. However, as we show in Section \ref{sec:resultsMain}, the use of a DNN alone does not provide high predictive performance because a DNN learns by considering features that are independent of each other. Conventional ML algorithms can identify how different types of data are interrelated and create new segments based on those relationships. To bring the best of both worlds together, we use clustering along with feature maps learned by DNN. This helps in finding the relationship between the features that can be segmented.

To efficiently generate binary masks, we need information about cloud boundaries. To obtain cloud boundaries, we apply \textit{k}-means   clustering on the RGB pixel values of the input image (refer \circled{F} in Figure \ref{fig:ModelArchitecture}). This clusters similar pixels together while creating a border between sky and cloud. We use a centroid value of two to create two clusters, viz., a cloud region, and a non-cloud region. 
We want to learn different features from cloud and sky from daytime and nighttime images. Hence, we have used \textit{k}-means and not edge detection algorithms like Sobel edge detector.

Following concatenation, the resultant feature map is processed through two Conv blocks having kernel size $3 \times 3$ to enhance the features (refer \circled{D} in Figure \ref{fig:ModelArchitecture}). Then, bilinear upsampling is performed.
The clustered input image is passed through Conv block with kernel size 1 (refer \circled{E} in Figure \ref{fig:ModelArchitecture}) and then added with the upsampled feature map obtained above. Finally, this feature map is passed through a $1 \times 1$ convolution layer, reducing the number of channels to two. This output is passed through sigmoid activation to generate the output mask.

Overall network design: Starting from \circled{B} in Fig. \ref{fig:ModelArchitecture}, we first up-sample the refined features. We concatenate this upsampled feature map with the matching low-level feature map from the network backbone, which has the same spatial resolution. Before this concatenation, we pass the low-level feature map obtained from the backbone network (refer \circled{C} in Figure \ref{fig:ModelArchitecture}) through a Conv block with kernal size of $1 \times 1$ to reduce the number of channels. This is because the low-level features typically contain a large number of channels, which can outweigh the importance of the output from ASPP features. 

Following concatenation, the resultant feature map is processed through two Conv blocks having kernal size $3 \times 3$ to enhance the features (refer \circled{D} in Figure \ref{fig:ModelArchitecture}). Then, bilinear upsampling is performed.
The clustered input image is passed through Conv block with a kernal size of $1 \times 1$ (refer \circled{E} in Figure \ref{fig:ModelArchitecture}) and then added with the upsampled feature map obtained above. Finally, this feature map is passed through a $1 \times 1$ convolution layer, reducing the number of channels to two. This output is passed through sigmoid activation to generate the output mask.

\section{Experimental Platform}\label{sec:experimentalplatform}
Dataset:
We have used SWINySEG \citep{dev2019cloudsegnet} dataset for training our model, which has images of size $300 \times 300$ pixels. SWINySEG is a composite dataset consisting of augmentations applied to the SWIMSEG \citep{dev2016color} and SWINSEG \citep{dev2017nighttime} dataset to create a balance between daytime and nighttime images. We use random sampling to divide the
training and testing sets in an 80:20 ratio on this composite dataset. The SWINySEG dataset has images consisting of a combination of images taken both during daytime and nighttime. The SWIMSEG dataset has 1013 images taken during the daytime, while on the other hand, SWINSEG has 115 images taken during the nighttime. The SWIMSEG dataset has 1013 images taken during the daytime, while on the other hand, SWINSEG has 115 images taken during the nighttime. During the inference phase, we resize the image to $300 \times 300$ pixels and then use a random crop.

 Training settings: We use TensorFlow 2.6. We use the Adam optimizer with an initial learning rate of 0.0001. Whenever there is no convergence for 20 epochs continuously, the learning rate is dynamically varied for fine-tuning.  We use a batch size of 8 to train the model.
We perform end-to-end training of ACLNet to minimize segmentation loss. For segmentation, we use BinaryCrossEntropy-DICE (BCE-DICE) loss.  BCE-DICE loss is a combination of the distribution-based loss function (BCE) and region-based loss function (DICE loss). BCE loss considers each pixel as an independent prediction and optimizes loss in under-segmented regions. DICE optimizes loss in over-segmented regions. The training is done for 300 epochs. The dice score saturates after 270 epochs.  Because loss is always convergent towards global minima, the network's training is consistent overall. We observe that use of \textit{k}-means clustering has little impact on the training and testing time of the network.

Evaluation metrics: We use (1) precision, (2) recall, (3) F1-score, and (4) error rate defined as $\frac{\text{FP+FN}}{\text{TP+FP+TN+FN}}$.  Here, FP means false positive, TP means true positive, FN means false negative and TN means true negative.   (5) mean intersection over union (MIoU), defined as  $\text{MIoU} = \frac{1}{\left | \text{C} \right |} \sum_{\text{c}\in \text{C}}^{} J_\text{c}(y^{*}, \tilde{y})$ 
where $y^{\star}$ and $\tilde{y}$ contain the ground truth and predicted labels of all pixels in the testing dataset.   C denotes all the classes and c denotes individual classes from C.  $J_\text{c}$ is the Jaccard index of class c.
 (6) ``Matthews correlation coefficient'' (MCC), defined as  
$\text{MCC} = \frac{\text{TP} \times \text{TN - FP} \times \text{FN}}{ \sqrt{ (\text{TP + FP})(\text{TP+FN})(\text{TN+FP})(\text{TN+FN})} }$. (7) We show the ROC (receiver operating characteristic)  curve, which  depicts classifier performance at all classification thresholds. 
We have performed three trials of experiments with ACLNet and observed that the standard deviation of results across different trials is negligibly small.

\section{Results}\label{sec:resultsMain}
\begin{figure}[ht]  \centering
\includegraphics [scale=0.50] {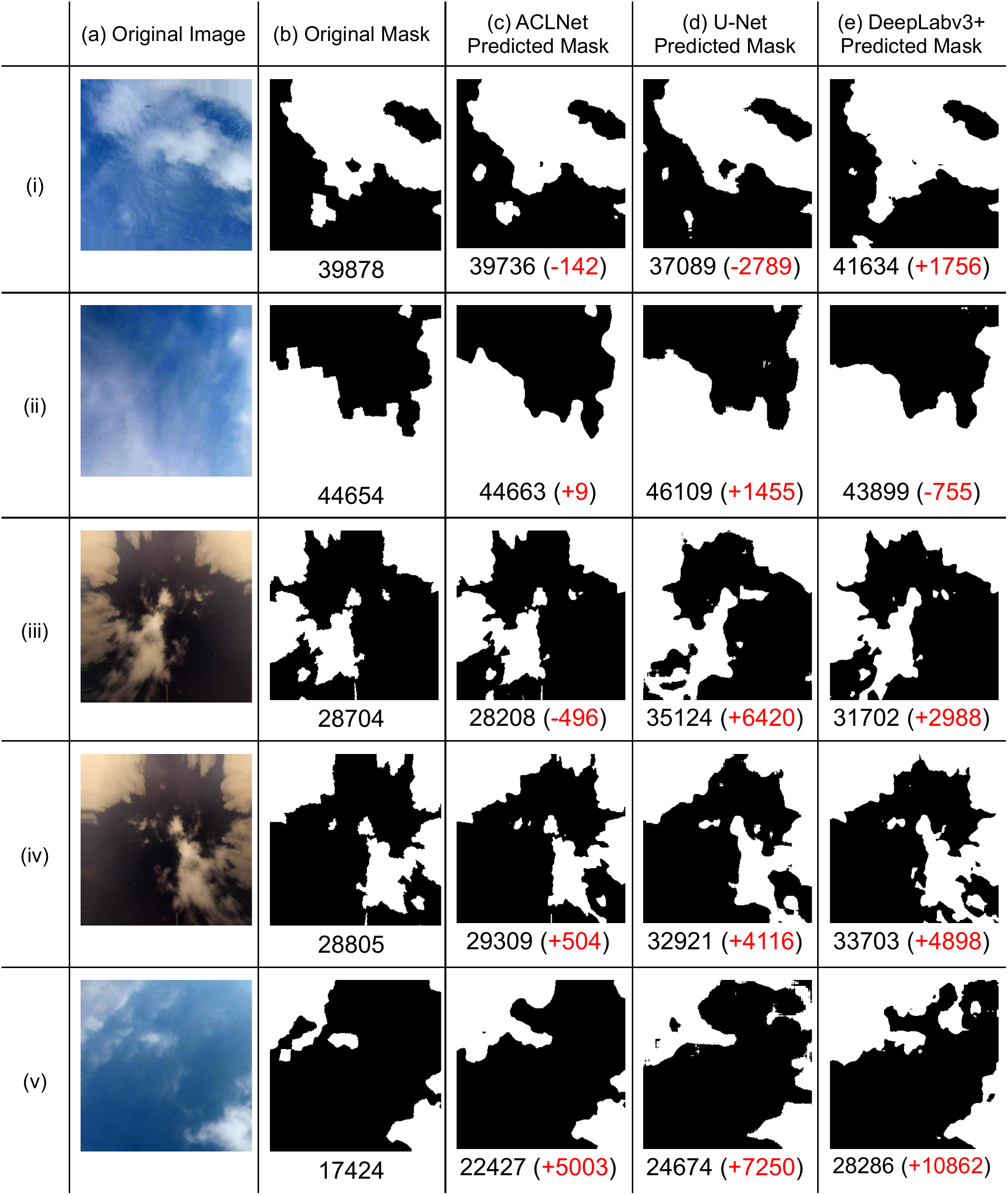}
\caption{Segmentation results for three daytime (i), (ii),(v) and two nighttime (iii)-(iv) images. The number of pixels in the cloud mask is shown below each figure. The red-color text in parenthesis shows the difference in pixel-count between the ground-truth and the prediction made by a model (ACLNet/U-Net/DeepLabV3+). }
 \label{fig:qualitativeresults}
  \end{figure}

Visualization: 
Figure \ref{fig:qualitativeresults} (i)-(v) shows five sample images (three daytime and two nighttime images) from the SWINySEG dataset. It also compares the ground truth masks (b) with the output masks produced by ACLNet (c), U-Net (d) and DeepLabv3+ (e). 
We can see that the output density map generated by ACLNet is quite similar to the ground-truth density estimation map.
ACLNet preserves the cloud boundaries, and by virtue of using the clustering, it preserves the pixel information for generating a binary mask. For all these images, ACLNet leads to more accurate pixel-count than U-Net and DeepLabv3+. In (v), none of the models give good predicted pixel count, still ACLNet performs much better than the other models.

Quantitative results:  We compare ACLNet with five other baseline networks, viz., FCN, CloudU-Net, DeepLabv3+, U-Net and CloudSegNet. Note that we have ourselves trained these networks and performed experiments to obtain the results.  Table \ref{tab:mainResults} shows the experimental results. Evidently, ACLNet provides the best value of metrics for all cases, except for precision metric on nighttime images.

\begin{table}[htbp]\footnotesize
\centering
\caption{ Results  on daytime, nighttime and day+night time images  (Best values are shown in \textbf{bold font}) }\label{tab:mainResults}
\begin{tabu}{|l|c|c|c|c|c|c|}
\hline
Method               & \multicolumn{1}{|c|}{Precision} & \multicolumn{1}{c|}{Recall} & \multicolumn{1}{c|}{F1-Score} & \multicolumn{1}{c|}{Error Rate} &  \multicolumn{1}{c|}{MIoU} &  \multicolumn{1}{l|}{MCC} \\\hline
\multicolumn{7}{|c|}{Daytime (SWIMSEG)}  \\\hline 
 FCN            &  	0.532		&	0.466		&	0.456		&	0.502			&	0.651		&		0.724     \\\hline
 CloudU-Net            &		0.951		&		0.971	&	0.952	&	0.042		&	0.963		&		0.853     \\\hline
 DeepLabv3+            & 	0.889	&	0.913	&	0.888	&	0.082			&	0.971	&		0.93              \\\hline
 U-Net               & 	0.771	&	0.772	&	0.754	&	0.191			
&	0.870		&	0.812        \\\hline
 CloudSegNet        & 	0.921	&	0.897	&	0.892	&	0.078				&	0.944		&	0.826  	    \\\hline
 CloudSegNet (with clustering)					& 	0.941		&	0.914	&	0.912	&	0.061			&	0.955			&	0.885  \\\hline
ACLNet (proposed)        & \textbf{0.964}                         & \textbf{0.979}   & \textbf{0.971}                        & \textbf{0.022}     & \textbf{0.992} 					&  \textbf{0.956}                   \\\hline
\multicolumn{7}{|c|}{Nighttime (SWINSEG)} \\\hline 
 FCN               &	 0.423		&	0.492	&	0.431	&	0.567	
&	0.591	&	0.681          \\\hline
 CloudU-Net        &   \textbf{0.943}	&	0.951	&	0.941	&	0.049	
&	0.931	&	0.816           \\\hline
 DeepLabv3+               & 	0.864	&	0.962	&	0.891	&	0.084			&	0.961	&	0.901        \\\hline
 U-Net               & 	0.693		&	0.673		&	0.701	&	0.240	
&	0.842	&	0.782	       \\\hline
 CloudSegNet       & 	0.891		&	0.924		&	0.881	&	0.083		&	0.915	&	0.824	        \\\hline
 CloudSegNet (with clustering)					&	0.932		&	0.934		&	0.901	&	0.075		&	0.931	&	0.861			\\\hline
ACLNet (proposed)        & 0.917                          & \textbf{0.982}                       & \textbf{0.947}                        &  \textbf{0.037}        & \textbf{0.985}    & \textbf{0.930}             \\\hline
\multicolumn{7}{|c|}{Day + Night Time (SWINySEG)} \\\hline 
 FCN                &   0.500		&		0.511		&		0.441	&	0.555		&	0.591			&		0.713   		\\\hline
 CloudU-Net           & 0.956		&	0.967		&	0.952	&	0.044	
&	0.941			&		0.945     		\\\hline
 DeepLabv3+                & 	0.861	&	0.906	&	0.852	&	0.084			&	0.973			&		0.93             \\\hline
 U-Net          & 	0.714		&	0.764	&	0.741	 &	0.216				&	0.855						&		0.8        						\\\hline
 CloudSegNet       & 	0.930	&	0.883	&	0.891	&	0.079				&	0.926			&		0.812     \\\hline
 CloudSegNet (with clustering)					&	0.931	& 	0.943	&	0.922	&	0.071				&	0.932			&	0.873		\\\hline
ACLNet (proposed)        &  \textbf{0.959 }                         & \textbf{0.979}               			& \textbf{0.968}     & \textbf{0.024} & \textbf{0.993} & \textbf{0.960} \\\hline                     
\end{tabu}
\end{table}

The error rate quantifies the rate of pixel-wise misclassification between ground truth and predicted output over the whole set of instances. Error rate measures the inaccuracy of predicted output values for target values. ACLNet achieves the lowest error-rate for all three categories. These results confirm the superiority of our model. 

Based on the histogram peak of daytime and nighttime images, the difference between the color of sky and cloud regions is better in daytime images than in nighttime images. Hence, ACLNet performs well on daytime images but produces slightly inferior results on nighttime images. Although ACLNet's precision on the nighttime images is slightly less, the cloud features are efficiently segmented. 

 In this paper, our key idea is that since machine-learning (ML) and deep-learning  (DL) networks extract complementary features, combining them can provide higher performance. Our ACLNet network validates this idea and for further validation, we evaluate combining \textit{k}-means clustering with CloudSegNet network. On comparing ``CloudSegNet'' results with ``CloudSegNet (with clustering)'' in Table \ref{tab:mainResults}, it is clear that \textit{k}-means clustering helps in improving the predictive performance of CloudSegNet also. This confirms our design-choice and the importance of our innovation.

\begin{figure}[htbp]  \centering
\includegraphics [scale=0.50] {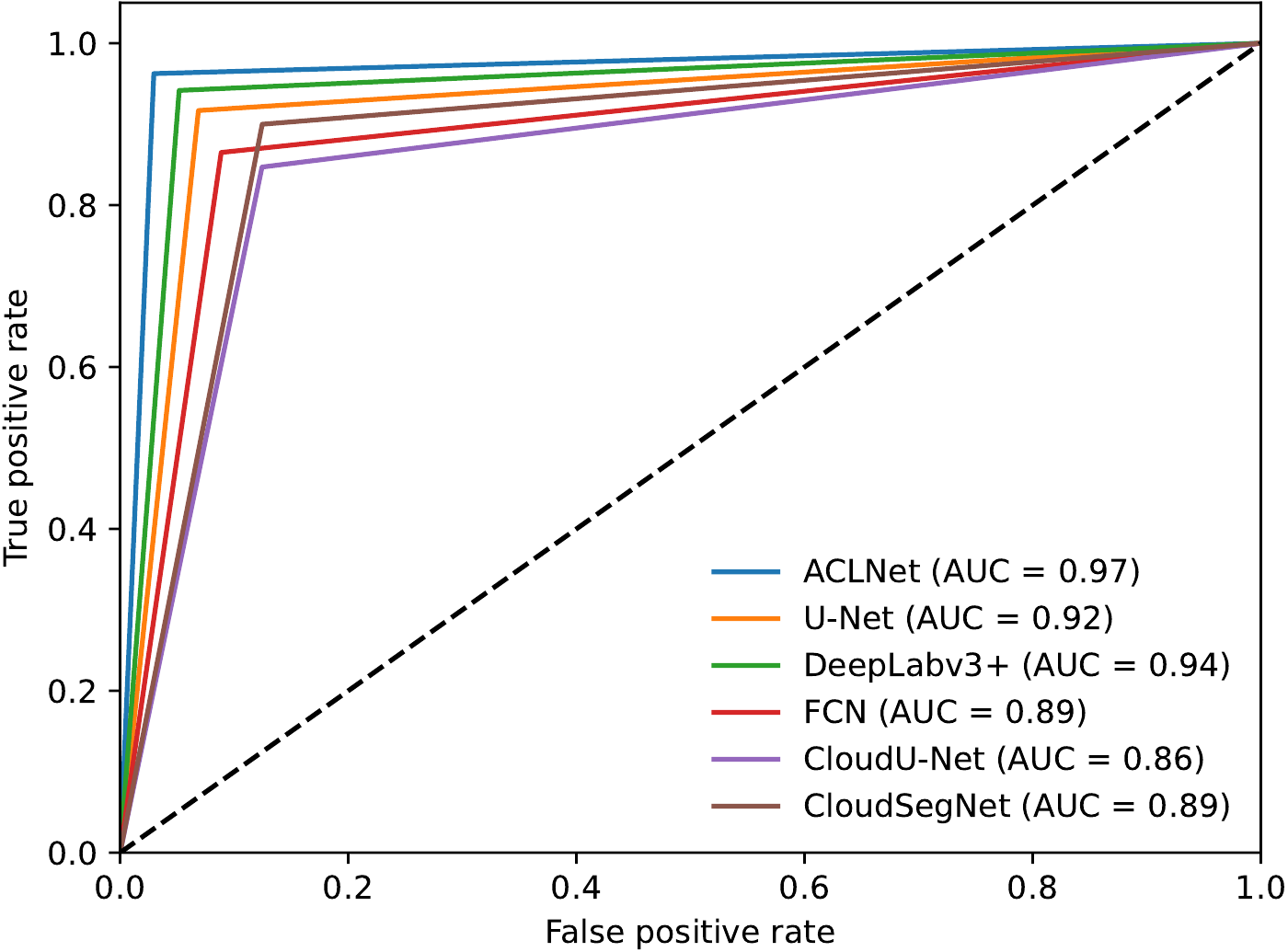}
\caption{ROC curve comparison of different segmentation models.}
 \label{fig:roccurve}
  \end{figure}

ROC curve: Figure \ref{fig:roccurve} compares the ROC curves of ACLNet with the other baseline networks shown in Table \ref{tab:mainResults}. The area under the curve (AUC) of ROC for CloudU-Net is the least while the AUC-ROC curve for ACLNet is the highest. The higher the true positive rate and lower the false positive rate, the higher is the AUC. This shows that ACLNet has the highest true positive rate while having the lowest false positive rate.

Model size and throughput:   
From Table \ref{tab:modelsize}, we can see that the model size of ACLNet is lower than all techniques, except CloudSegNet. While  CloudSegNet has a small model size,  but its predictive performance is much inferior compared to ACLNet. CloudU-Net has a very high model size (138MB).  Overall, ACLNet achieves the best balance of model size and performance.
Also, ACLNet requires only 7.57 million FLOPs. 
 ACLNet has a throughput of 5.7 and  13.1 frames-per-second on 2080Ti GPU and  P100 GPU, respectively.  
 
\begin{table}[htbp]\footnotesize
  \centering
  \caption{Model size (MB) of different cloud segmentation models}
    \begin{tabular}{|c|c||c|c|}
    \hline
    Model &  Size  & Model &  Size \\
    \hline
    CloudU-Net & 138.6 & CloudSegNet & 0.11\\
    \hline
    DeepLabv3+ & 61.3  & ACLNet with EfficientNet-B1 backbone & 36.7\\
    \hline
    U-Net & 94.3  & ACLNet with EfficientNet-B2 backbone & 43.1\\
    \hline
    FCN \citep{long2015fully}  & 53.6  & ACLNet with ViT backbone & 325\\
    \hline
    ACLNet with ResNet50 backbone & 47.1  & ACLNet & 30.79\\
    \hline
    \end{tabular}%
  \label{tab:modelsize}%
\end{table}%

\textbf{Ablation studies:}

1. Impact of removing GAM and \textit{k}-means: We now evaluate the contribution of GAM and \textit{k}-means  by removing them individually and together. Table \ref{tab:GAMkmeans} shows the results. In Figure \ref{fig:ModelArchitecture}, on removing GAM, the path shown with \circled{A}-\circled{B} is removed. On removing \textit{k}-means clustering, the path shown with \circled{F}-\circled{E} in Figure \ref{fig:ModelArchitecture} is removed. On removing the \textit{k}-means clustering, all metrics become worse. Clearly, the features extracted by DNN are not sufficient for achieving high predictive performance for cloud segmentation. On removing the GAM, all metrics, especially the precision for nighttime images, become worse. ACLNet without \textit{k}-means gives inferior results as compared to ACLNet without GAM in precision and F1-Score for nighttime images, whereas the results are almost equal for the daytime and day+night images.
On removing both \textit{k}-means and GAM, the results become even worse (results omitted for brevity). Thus, both \textit{k}-means and attention modules are important for achieving high performance.

\begin{table}[htbp]
  \centering
  \caption{Results of ablation studies on GAM and \textit{k}-means clustering  (P=precision, R=recall, ER=error-rate) }
   \begin{adjustbox}{width=1\textwidth}
    \begin{tabular}{|c|c|c|c|c|c|c|c|c|c|c|c|c|}
    \hline
               & \multicolumn{6}{c|}{ACLNet without GAM} & \multicolumn{6}{c|}{ACLNet without \textit{k}-means} \\
    \hline
        & P & R & F1 & ER & MIoU & MCC & P & R & F1 & ER & MIoU & MCC \\
    \hline
    Day      & 0.94   & 0.93   & 0.94   & 0.05   & 0.97   & 0.91   & 0.92   & 0.95   & 0.94   & 0.073  & 0.96   & 0.918 \\
    \hline
    Night    & 0.82   & 0.97   & 0.88   & 0.078  & 0.95   & 0.87   & 0.91   & 0.92   & 0.93   & 0.078  & 0.95   & 0.895 \\
    \hline
    Day + night   & 0.94   & 0.92   & 0.94   & 0.05   & 0.98   & 0.9    & 0.93   & 0.94   & 0.93   & 0.074  & 0.96   & 0.913  \\ \hline
    \end{tabular}%
    \end{adjustbox}
  \label{tab:GAMkmeans}%
\end{table}%

2. Impact of changing the backbone:  
Table \ref{tab:ablationbackboneResults} shows the results of replacing the backbone in ACLNet to four other networks: EfficientNet-B1 and B2, ViT and ResNet50. We observe that for all these four backbones, the predictive performance does not increase much compared to that with EfficientNet-B0. Further, on using ViT, the model size increases to 325 MB. This is much larger than the model size with EfficientNet-B0 backbone. 
On 2080Ti GPU and P100 GPU, ACLNet (with EfficientNet-B0 backbone) have throughput of 5.7 and 13.1 frames-per-second, respectively. For ACLNet (with ViT backbone), these numbers are 1.8 and 5.0, respectively. Thus,  use of EfficientNet-B0 leads to higher throughput than use of ViT. 
Considering predictive performance, model size and throughput, we prefer EfficientNet-B0 as the backbone.

\begin{table}[htbp]\footnotesize
  \centering 
  \caption{Results of ablation studies on ACLNet with different backbones (P=precision, R=recall, ER=error-rate)}\label{tab:ablationbackboneResults}%
    \begin{tabular}{|c|c|c|c|c|c|c|c|c|c|c|c|c|}
    \hline
      & \multicolumn{6}{c|}{ACLNet with EfficientNet-B1 backbone} & \multicolumn{6}{c|}{ACLNet with EfficientNet-B2 backbone} \\
    \hline
      & P & R & F1 & ER & MIoU & MCC & P & R & F1 & ER & MIoU & MCC \\
    \hline
    Day    & 0.96   & 0.98   & 0.97   & 0.022  & 0.98   & 0.95   & 0.98  & 0.98  & 0.98  & 0.022  & 0.98  & 0.95 \\
    \hline
    Night  & 0.91   & 0.98   & 0.95   & 0.033  & 0.98   & 0.94   & 0.92  & 0.98  & 0.96  & 0.033  & 0.97  & 0.94 \\
    \hline
    Day + night & 0.96   & 0.98   & 0.97   & 0.023  & 0.97   & 0.95   & 0.96  & 0.98  & 0.98  & 0.022  & 0.99  & 0.95 \\
    \hline \hline
      & \multicolumn{6}{c|}{ACLNet with ViT backbone} & \multicolumn{6}{c|}{ACLNet with ResNet50 backbone}    \\
    \hline    
    Day    & 0.98   & 0.98   & 0.98   & 0.022  & 0.99   & 0.95   & 0.96   & 0.98   & 0.96   & 0.025  & 0.98   & 0.95 \\
    \hline
    Night  & 0.94   & 0.97   & 0.95   & 0.031  & 0.98   & 0.94   & 0.92   & 0.98   & 0.95   & 0.037  & 0.96   & 0.93 \\
    \hline
    Day + night & 0.97   & 0.98   & 0.97   & 0.022  & 0.99   & 0.95   & 0.96   & 0.98   & 0.97   & 0.026  & 0.98   & 0.95 \\
    \hline
    \end{tabular}% 
\end{table}%

3. Use of \textit{k}-means++  clustering: \textit{k}-means++ \citep{arthur2006k} acts as an intelligent initialization procedure for \textit{k}-means. 
\textit{k}-means++ algorithm minimizes the likelihood of a wrong initialization but has higher computational overhead. However, we observe that \textit{k}-means++ algorithm provides no improvement over the \textit{k}-means algorithm. 
This is because of the use of a smaller K value (two), such that random or intelligent initialization has a negligible impact.

\section{Conclusion and  Future Work}\label{sec:conclusion}

We present a novel model, ACLNet, that can accurately segment clouds from both nocturnal and daytime images. ACLNet achieves high predictive performance by virtue of combining the benefits of ``\`a trous spatial pyramid pooling'', attention and clustering. ACLNet is lightweight and has the highest recall and F1-score for all types of images.

 ACLNet is trained to work on the predetermined classes - namely cloud and sky.  However, a real-world image may also have other objects such as birds or plane. ACLNet may not work well with such unseen classes. To achieve better performance on those unseen classes, we would need to retrain or fine-tune our model. In near future, we plan to train ACLNet using unsupervised learning to further improve its segmentation performance. Vision transformer models provide state-of-art results on several computer vision tasks such as classification \citep{liu2021swin} and segmentation \citep{zheng2021rethinking,xie2021segformer,strudel2021segmenter}. Our future work will focus on comprehensively evaluating transformer-based models for cloud segmentation.

{\footnotesize
\bibliographystyle{tfcad}
\bibliography{References}
}

\end{document}